\def\year{2021}\relax
\documentclass[letterpaper]{article} % DO NOT CHANGE THIS
\usepackage{aaai21}  % DO NOT CHANGE THIS
\usepackage{times}  % DO NOT CHANGE THIS
\usepackage{helvet} % DO NOT CHANGE THIS
\usepackage{courier}  % DO NOT CHANGE THIS
\usepackage[hyphens]{url}  % DO NOT CHANGE THIS
\usepackage{graphicx} % DO NOT CHANGE THIS
\usepackage{natbib}  % DO NOT CHANGE THIS AND DO NOT ADD ANY OPTIONS TO IT
\usepackage{caption} % DO NOT CHANGE THIS AND DO NOT ADD ANY OPTIONS TO IT
\usepackage{graphicx}
\usepackage{amsmath,amsthm,amssymb,amsfonts}
\usepackage{booktabs}
\usepackage{multirow}
\usepackage{enumitem}

\usepackage{CJKutf8}
\newcommand{\songti}[1]{\begin{CJK*}{UTF8}{gbsn} #1  \end{CJK*}}

\title{Generating Diversified Comments via Reader-Aware Topic Modeling and\\ Saliency Detection}

\author{Wei Wang$^{1,2,}$\thanks{Work was done during internship at Tencent AI Lab.}~, Piji Li$^3$, Hai-Tao Zheng$^{1,2,}$\thanks{Corresponding author.} \\
$^1$Shenzhen International Graduate School, Tsinghua University \\ 
$^2$Department of Computer Science and Technology, Tsinghua University \\
$^3$Tencent AI Lab \\
 {\tt w-w16@mails.tsinghua.edu.cn,pijili@tencent.com} \\
   {\tt zheng.haitao@sz.tsinghua.edu.cn}
}

\begin{document}
\maketitle
\begin{abstract}
Automatic comment generation is a special and challenging task to verify the model ability on news content comprehension and language generation.
Comments not only convey salient and interesting information in news articles, but also imply various and different reader characteristics which we treat as the essential clues for \textbf{diversity}.
However, most of the comment generation approaches only focus on saliency information extraction, while the reader-aware factors implied by comments are neglected.
To address this issue, we propose a unified reader-aware topic modeling and saliency information detection framework to enhance the quality of generated comments.
For reader-aware topic modeling, we design a variational generative clustering algorithm for latent semantic learning and topic mining from reader comments.
For saliency information detection, we introduce Bernoulli distribution estimating on news content to select saliency information. The obtained topic representations as well as the selected saliency information are incorporated into the decoder to generate diversified and informative comments.
Experimental results on three datasets show that our framework outperforms existing baseline methods in terms of both automatic metrics and human evaluation. The potential ethical issues are also discussed in detail.
\end{abstract}

\section{Introduction}

% In this paper, we focus on the task of automatic news comment generation. With the news portals and Apps becoming more and more popular, users can easily touch the news feed, pay attention to the information they are interested in, and express their opinions via a great diversity of comments~\cite{tsagkias2010news,ksiazek2018commenting}. Automatic comment generation system helps user produce comment drafts and improves the user participation to the news products. 
For natural language generation research, automatic comment generation is a challenging task to verify the model ability on aspects of news information comprehension and high-quality comment generation~\cite{reiter1997building, gatt2018survey,qin-etal-2018-automatic,huang2020generating}.

\begin{figure}[!t]
\renewcommand\arraystretch{0.9}
% \resizebox{1\columnwidth}{!}{
\begin{tabular}{p{0.95\columnwidth}} 
\toprule
\small \textbf{Title:}   Andrew Lincoln Poised To Walk Off ‘The Walking Dead’ Next Season; Norman Reedus To Stay \\ \midrule
\small \textbf{Body   (truncated):} He has lost his on-screen son, his wife and a number of friends   to the zombie apocalypse, and now The Walking Dead star Andrew Lincoln looks   to be taking his own leave of the AMC blockbuster. Cast moves on the series   will also see Norman Reedus stay put under a new \$ 20 million contract.  \\      
\small Almost unbelievable on a show where almost no one is said to be safe, the man who has played Rick Grimes could be gone by the end of the upcoming ninth season... The show will reset with Reedus ’ Daryl Dixon even more in the spotlight ... \\ \midrule
\small \textbf{Comment A:} I'm not watching TWD without Lincoln.    \\  
\small \textbf{Comment B:} Storylines getting stale and they keep having the same type trouble every year. \\
\small \textbf{Comment C:} Reedus ca n't carry the show playing Daryl.   \\
\bottomrule                                     
\end{tabular}
% }
    \caption{A news example from Yahoo.}
    \label{yahoo_case}
	\vspace{-8mm}
\end{figure}

One common phenomenon is that, for the same news article, there are usually hundreds or even thousands of different comments proposed by readers in different backgrounds. Figure~\ref{yahoo_case} depicts an example of a news article (truncated) and three corresponding comments from Yahoo. The news article is about ``The Walking Dead''. From the example, we can observe and conclude two characteristics: (1) Although the news article depicts many different aspects of the event and conveys lots of detailed information, readers usually pay attention to part of the content, which means that not all the content information is \textbf{salient} and \textbf{important}. As shown in Figure \ref{yahoo_case}, the first two readers both focus on ``The Walking Dead'' and the third reader focuses on ``Reedus''.  None of them mention other details in the content. (2) Different readers are usually interested in different topics, and even on the same topic, they may hold different opinions, which makes the comments \textbf{diverse} and \textbf{informative}. For example, the first reader gives the comment from the topic of ``feeling'' and he ``cannot accept Lincoln's leaving''. The second reader comments on the topic of ``plot'' and thinks ``it is old-fashioned''. The third reader comments from the topic of ``acting'', saying that ``Reedus can't play that role''.  

Therefore, news comments are produced based on the interactions between readers and news articles. Comments not only imply different important information in news articles, but also convey distinct reader characteristics. Precisely because of these reader-aware factors, we can obtain a variety of diversified comments under the same news article.
Intuitively, as the essential reason for diversity, these reader-aware factors should be considered jointly with saliency information detection in the task of diversified comment generation. 
% This inspires us to consider the diversity of user topics in this task. such as the special writing styles and contrary opinions on the same event.   
% However, rare of works concern these two components jointly in the proposed comment generation frameworks.
% Several previous studies have been conducted on news comment generation. It is not suitable to use the seq2seq framework directly.
However, it is rare that works consider these two components simultaneously. \citeauthor{zheng2017automatic}\shortcite{zheng2017automatic} proposed to generate one comment only based on the news title. \citeauthor{qin-etal-2018-automatic}\shortcite{qin-etal-2018-automatic} extended the work to generate a comment jointly considering the news title and body content. These two seq2seq~\cite{sutskever2014sequence} based methods conducted saliency detection directly via attention modeling. \citeauthor{li-etal-2019-coherent}\shortcite{li-etal-2019-coherent} extracted keywords as saliency information. \citeauthor{yang-etal-2019-read}\shortcite{yang-etal-2019-read} proposed a reading network to select important spans from the news article. \citeauthor{huang2020generating}\shortcite{huang2020generating} employed the LDA \cite{blei2003latent} topic model to conduct information mining from the content. All these works concern the saliency information extraction and enhance the quality of comments. However, various reader-aware factors implied in the comments, which are the essential factors for diversity as well, are neglected.   

% These methods solve it as a one-to-one mapping problem, which can not effectively model this one-to-many relationship. In addition, it is necessary to select important information of news for comment generation. However, extracting keywords as important information  \cite{li-etal-2019-coherent} can not dynamically select important information based on comments. The reading network \cite{yang-etal-2019-read} can dynamically select important spans but it needs reinforce learning to train the model, which may lead to instability in training. These inspire us to design a content selection mechanism that can dynamically select important information and is easy to train.

To tackle the pre-mentioned issues, we propose a reader-aware topic modeling and saliency detection framework to enhance the quality of generated comments.
The goal of reader-aware topic modeling is to conduct reader-aware latent factors mining from the \textbf{comments}. The latent factors might be either reader interested topics or the writing styles of comments, or more other detailed factors. We do not design a strategy to disentangle them, instead, we design a unified latent variable modeling component to capture them. Specifically, inspired by \citet{jiang2017variational}, we design a Variational Generative Clustering (VGC) model to conduct latent factors modeling from the reader comments. The obtained latent factor representations can be interpreted as news topics, user interests, or writing styles. For convenience, we collectively name them as \textbf{Topic}.
For reader-aware saliency information detection, we build a saliency detection component to conduct the Bernoulli distribution estimating on the news content. Gumbel-Softmax is introduced to address the  non-differentiable sampling operation. Finally, the obtained topic representation vectors and the selected saliency news content are integrated into the generation framework to control the model to generate diversified and informative comments.
%In addition, a saliency detection module is designed to select important information of news. It predicts a label 0 or 1 for each word in the news, which indicates whether it it important or not. Only words with label 1 are preserved for comment generation. The comment generation component generates a comment conditioning on the selected content and the selected topic. 
% Specifically, the reader-aware topic modeling component is consisted of a comment reconstructor and a topic classifier. The comment reconstructor adopts the structure similar to VAE \cite{kingma2014auto} and is used to learn latent representations of comments. The topic classifier classify comments into certain topics. The comment generation component builds on seq2seq framework and a saliency information detector is inserted between the news encoder and the comment decoder. 

We conduct extensive  experiments on three datasets in different languages: NetEase News (Chinese)~\cite{zheng2017automatic}, Tencent News (Chinese)~\cite{qin-etal-2018-automatic}, and Yahoo! News (English)~\cite{yang-etal-2019-read}. Experimental results demonstrate that our model can obtain better performance according to automatic evaluation and human evaluation. In summary, our contributions are as follows:
\begin{itemize}
    \setlength\itemsep{-0.2em}
    \item We propose a framework to generate diversified comments jointly considering saliency news information detection and reader-aware latent topic modeling.
    \item We design a Variational Generative Clustering (VGC) based component to learn the reader-aware latent topic representations from comments.
    \item For reader-aware saliency information detection, Bernoulli distribution estimating is conducted on the news content. Gumbel-Softmax is introduced to address the non-differentiable sampling operation.
    \item Experiments on three datasets demonstrate that our model outperforms state-of-the-art baseline methods according to automatic evaluation and human evaluation.
\end{itemize}

\section{Methodology}

\subsection{Overview}

\begin{figure*}[!t]
\centering
\includegraphics[width=16cm]{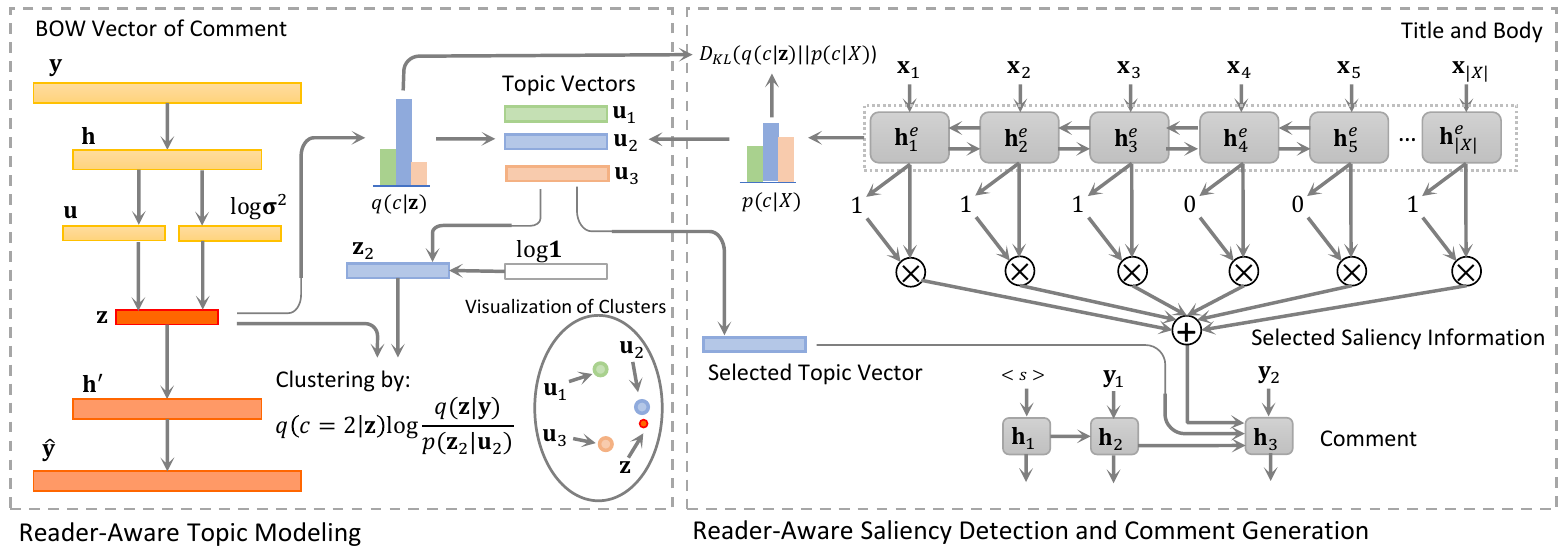}
\caption{The framework of our proposed method. Left: the reader-aware topic modeling component is used to learn topic vectors. Specifically, the comment $\mathbf{y}$ is encoded to get latent semantic vector $\mathbf{z}$. Then the classifier $q(c|\mathbf{z})$ classifies $\mathbf{z}$ into one topic. In addition, $\mathbf{z}$ is decoded to reconstruct the comment. Topic vectors are learned according to Equation \ref{cluster_loss}. Right: the reader-aware saliency detection is used to select saliency words and the topic selector $p(c|X)$ is used to select an appreciate topic vector. Finally the comment is generated conditioning on the selected topic vector and selected saliency information.}
\label{all_method}
\vspace{-5mm}
\end{figure*}

To begin with, we state the problem of news comment generation as follows: given a news title $T = \{t_1,t_2,\ldots,t_m\}$ and a news body $B=\{b_1,b_2,\ldots,b_n\}$, the model needs to generate a comment $Y = \{y_1,y_2,\ldots,y_l\}$ by maximizing the conditional probability $p(Y|X)$, where $X = [T, B]$. 
As shown in Figure \ref{all_method}, the backbone of our work is a sequence-to-sequence framework with attention mechanism \cite{bahdanau2014neural}. Two new components of reader-aware topic modeling and saliency information detection are designed and incorporated for better comment generation.
For reader-aware topic modeling, we design a variational generative clustering algorithm for latent semantic learning and topic mining from reader comments.
For reader-aware saliency information detection, we introduce a saliency detection component to conduct the Bernoulli distribution estimating on news content.
Finally, the obtained topic vectors as well as the selected saliency information are incorporated into the decoder to conduct comment generation.

\subsection{The Backbone Framework}
\label{encattdec}
The backbone of our work is a sequence-to-sequence framework with attention mechanism \cite{bahdanau2014neural}.
A BiLSTM encoder is used to encode the input content words $X$ into vectors. Then a LSTM-based decoder generates a comment $Y$ conditioning on a weighted content vector computed by attention mechanism. Precisely, a word embedding matrix is used to convert content words into embedding vectors. Then these embedding vectors are fed into encoder to compute the forward hidden vectors via:
\begin{equation}
\setlength{\abovedisplayskip}{3pt}
\setlength{\belowdisplayskip}{3pt}
\overrightarrow{\mathbf{h}_i} = LSTM_{f}^{e} \left(\mathbf{x} _i, \overrightarrow{\mathbf{h}}_{i-1} \right),
\end{equation}
where $\overrightarrow{\mathbf{h}_i} $ is a $d$-dimension hidden vector and $LSTM^e_f$ denotes the LSTM unit. The reversed sequence is fed into $LSTM^e_b$ to get the backward hidden vectors. We concatenate them to get the final hidden representations of content words. And the representation of news is : $\mathbf{h}^e=\left [\overrightarrow{\mathbf{h}_{|X|}};\overleftarrow{\mathbf{h}_0} \right]$.

The state of decoder is initialized with $\mathbf{h}^e$. For predicting comment word $y_t$, the hidden state is first obtained by:
\begin{equation}
\setlength{\abovedisplayskip}{3pt}
\setlength{\belowdisplayskip}{3pt}
\label{hupdate}
\mathbf{h} _ { t } = LSTM_d \left(  \mathbf{y}_ { t - 1 }, \mathbf{h} _ { t - 1 } \right),
\end{equation}
where $h_t \in \mathbb{R}^{2d}$ is the hidden vector of comment word and $LSTM_d$ is the LSTM unit. Then we use attention mechanism to query the content information from source input. The weight of each content word is computed as follows:
\begin{equation}
\begin{aligned}
e_{ti} &= \mathbf{h}_t^\top W_a \mathbf{h}^e_i ,\\
\alpha _{ti} &= \frac { \exp \left( e_{ti} \right) } { \sum _{i{\prime=1}}^{|X|} \exp \left( e_{ti{\prime}} \right) },
\end{aligned}
\end{equation}
where $W_a \in \mathbb{R}^{2d \times 2d}$, $\mathbf{h}^e_i$ is the hidden vector of content word $i$, $\alpha_{ti}$ is the normalized attention score on $x_i$ at time step $t$. Then the attention-based content vector is obtained by:
$
\tilde{\mathbf{h}}^e _ { t } = \sum_i \alpha _ { t i } \mathbf{h}^e _ { i }
$. The hidden state vector is updated by:
\begin{equation}
\setlength{\abovedisplayskip}{3pt}
\setlength{\belowdisplayskip}{3pt}
  \tilde{\mathbf{h}}_t = \operatorname{tanh}(W_c[\mathbf{h}_t;\tilde{\mathbf{h}}^e _ { t }]).
\end{equation}
Finally, the probability of the next word $y_t$ is computed via:
\begin{equation}
\setlength{\abovedisplayskip}{3pt}
\setlength{\belowdisplayskip}{3pt}
\label{predict}
p \left( y_ { t } | y_ { t - 1 } , X \right) = \operatorname { softmax } \left( linear \left( \tilde{\mathbf{h}} _ { t } \right) \right),
\end{equation}
where $linear(\cdot)$ is a linear transformation function.

During training, cross-entropy loss $\mathcal{L}_{ce}$ is employed as the optimization objective.

\subsection{Reader-Aware Topic Modeling}
Reader-aware topic modeling is conducted on all the comment sentences, aiming to learn the reader-aware topic representations only from the comments. To achieve this goal, we design a variational generative clustering algorithm which can be trained jointly with the whole framework in an end-to-end manner.

Since this component is a generative model, thus for each comment sentence $Y$ (we employ Bag-of-Words feature vector $\mathbf{y}$ to represent each comment sentence), the generation process is: (1) A topic $c$ is generated from the prior topic categorical distribution $p(c)$; (2) A latent semantic vector $\mathbf{z}$ is generated conditionally from a Gaussian distribution $p(\mathbf{z}|c)$; (3) $\mathbf{y}$ is generated from the conditional distribution $p(\mathbf{y}|\mathbf{z})$. According to the generative process above, the joint probability $p(\mathbf{y}, \mathbf{z}, c)$ can be factorized as:
\begin{equation}
\label{decode_process}
    p(\mathbf{y}, \mathbf{z}, c)=p(\mathbf{y} | \mathbf{z}) p(\mathbf{z} | c) p(c).
\end{equation}

By using Jensen’s inequality, the log-likelihood can be written as:
\begin{equation}
\small
\begin{aligned}
\log p(\mathbf{y}) &=\log \int_{\mathbf{z}} \sum_{c} p(\mathbf{y}, \mathbf{z}, c) d\mathbf{z} \\
& \geq E_{q(\mathbf{z}, c| \mathbf{y})}\left[\log \frac{p(\mathbf{y}, \mathbf{z}, c)}{q(\mathbf{z}, c | \mathbf{y})}\right]=\mathcal{L}_{\mathrm{ELBO}}(\mathbf{y}),
\end{aligned}\end{equation}
where $\mathcal{L}_{\mathrm{ELBO}}$ is the evidence lower bound (ELBO), $q(\mathbf{z}, c| \mathbf{y})$ is the variational posterior to approximate the true posterior $p(\mathbf{z}, c| \mathbf{y})$ and can be factorized as follows:
\begin{equation}
\label{encode_process}
q(\mathbf{z}, c | \mathbf{y})=q(\mathbf{z} | \mathbf{y}) q(c | \mathbf{z}).
\end{equation}
Then based on Equation \ref{decode_process} and Equation \ref{encode_process}, the $\mathcal{L}_{\mathrm{ELBO}}$ can be rewritten as:
\begin{equation}
\setlength{\abovedisplayskip}{3pt}
\setlength{\belowdisplayskip}{3pt}
\label{cluster_loss}
\small
\begin{aligned}
\mathcal{L}_{\mathrm{ELBO}}(\mathbf{y})=& E_{q(\mathbf{z}, c | \mathbf{y})}\left[\log \frac{p(\mathbf{y}, \mathbf{z}, c)}{q(\mathbf{z}, c | \mathbf{y})}\right] \\
=& E_{q(\mathbf{z}, c | \mathbf{y})}[\log p(\mathbf{y}, \mathbf{z}, c)-\log q(\mathbf{z}, c | \mathbf{y})] \\
=& E_{q(\mathbf{z}, c | \mathbf{y})}[\log p(\mathbf{y} | \mathbf{z})+\log p(\mathbf{z} | c)\\
&+\log p(c)-\log q(\mathbf{z} | \mathbf{y})-\log q(c | \mathbf{z})] \\
=& E_{q(\mathbf{z}, c | \mathbf{y})}[\log p(\mathbf{y} | \mathbf{z})\\
&- \log \frac{q(\mathbf{z}|\mathbf{y})}{p(\mathbf{z}|c)} - \log \frac{q(c|\mathbf{z})}{p(c)}] \\
=& E_{q(\mathbf{z}| \mathbf{y})}[\log p(\mathbf{y} | \mathbf{z})\\
&- \sum_{c}q(c | \mathbf{z})\log \frac{q(\mathbf{z}|\mathbf{y})}{p(\mathbf{z}|c)} \\
&-  D_{K L}(q(c | \mathbf{z})|| p(c)) ]\\
\end{aligned}
\end{equation}
where the first term in Equation \ref{cluster_loss} is the reconstruction term, which encourages the model to reconstruct the input. The second term aligns the latent vector $\mathbf{z}$ of input $\mathbf{y}$ to the latent topic representation corresponding to topic $c$. $q(c | \mathbf{z})$ can be regarded as the clustering result for the input comment sentence $Y$. The last term is used to narrow the distance between posterior topic distribution $q(c|\mathbf{z})$ and prior topic distribution $p(c)$.
%Next we give the specific implementation of each formula in Equation \ref{decode_process} and Equation \ref{encode_process}.

In practical implementations, the prior topic categorical distribution $p(c)$ is set to uniform distribution $p(c)=\frac{1}{K}$ to prevent the posterior topic distribution $q(c|\mathbf{z})$ from collapsing, that is, all comments are clustered into one topic. $p(\mathbf{z}|c)$ is a parameterised diagonal Gaussian as follows:
\begin{equation}
\setlength{\abovedisplayskip}{3pt}
\setlength{\belowdisplayskip}{3pt}
    p(\mathbf{z}|c) = \mathcal{N}\left(\mathbf{z} | \boldsymbol{\mu_c}, \operatorname{diag}\left(\boldsymbol{1}\right)\right),
\end{equation}
where $\boldsymbol{\mu_c}$ is mean of Gaussian for topic $c$, which is also used as the latent topic representation of topic $c$. Inspired by VAE \cite{kingma2014auto,bowman2016generating}, we employ a parameterised diagonal Gaussian as $q(\mathbf{z}|\mathbf{y})$:
\begin{equation}
\setlength{\abovedisplayskip}{3pt}
\setlength{\belowdisplayskip}{3pt}
\begin{aligned}
    \boldsymbol{\mu}&=l_{1}(\mathbf{h}), \log \boldsymbol{\sigma}=l_{2}(\mathbf{h}), \\
  q(\mathbf{z}|\mathbf{y})&=\mathcal{N}\left(\mathbf{z} | \boldsymbol{\mu}, \operatorname{diag}\left(\boldsymbol{\sigma}^{2}\right)\right),
\end{aligned}
\end{equation}
where $l_1(\cdot)$ and $l_2(\cdot)$ are linear transformations, $\mathbf{h}$ is obtained by the comment encoder, which contains two MLP layers with $\operatorname{tanh}$ activation functions. In addition, a classifier with two MLP layers is used to predict topic distribution $q(c|\mathbf{z})$. $p(\mathbf{y} | \mathbf{z})$ is modeled by the decoder, which is a one-layer MLP with $\operatorname{softmax}$ activation function.

After training, $K$ reader-aware topic representation vectors $\{\boldsymbol{\mu}_i\}_{1}^{K}$ are obtained only from the whole reader comments corpus in the training set. And these reader-aware topics can be used to control the topic diversity of the generated comments.

\subsection{Reader-Aware Saliency Detection}
% The news article is relatively long while the comments are relatively short and often focus on part of the news article. So there are a lot of information irrelevant to comments in the news article. Thus 
Reader-aware saliency detection component is designed to select the most important and reader-interested information from the news article. It conducts Bernoulli distribution estimating on each word of the news content, which indicates whether each content word is important or not. Then we can preserve the selected important words for comment generation.

Specifically, the saliency detection component first uses a BiLSTM encoder to encode title words and the last hidden vectors of two directions are used as the title representation $\mathbf{h}^{te}$. Then we use two-layer MLP with $\operatorname{sigmoid}$ activation function on the final layer to predict the selection probability for each content word $x_i$ as follows jointly considering the title information:
\begin{equation}
\setlength{\abovedisplayskip}{3pt}
\setlength{\belowdisplayskip}{3pt}
  p_\theta(\beta_i|x_i) = \operatorname{MLP}(\mathbf{h}^e_i, \mathbf{h}^{te}),
\end{equation}
where $\mathbf{h}^e_i$ is the hidden vector obtained by BiLSTM-based news content encoder in Section \ref{encattdec} 2.2. The probability $\beta_i$ determines the probability (saliency) of the word be selected, and it is used to parameterize a Bernoulli distribution. Then a binary gate for each word can be obtained by sampling from the Bernoulli distribution:
\begin{equation}
\label{eq:sampling}
\setlength{\abovedisplayskip}{3pt}
\setlength{\belowdisplayskip}{3pt}
  g_i \sim Bernoulli \left(\beta_i \right)
\end{equation}
Content words with $g_i=1$ are selected as context information to conduct the comment generation. Thus, the weight of each content word in attention mechanism and the weighted source context in the backbone framework in Section~\ref{encattdec} 2.2 are changed as follows:
\begin{equation}
\setlength{\abovedisplayskip}{3pt}
\setlength{\belowdisplayskip}{3pt}
\begin{aligned}
  \hat{\alpha}_{ti}&=\frac{g_i \odot \exp (e_{ti})}{\sum_{i\prime=1}^{|X|} g_{i{\prime}} \odot \exp (e_{ti^{\prime}})} ,\\
  \tilde{\mathbf{h}}^e _ { t } &= \sum _i \hat{\alpha} _ { t i } \mathbf{h}^e_i ,
\end{aligned}
\end{equation}
where $\tilde{\mathbf{h}}^e _ { t }$ is the selected saliency information of news content and it will be used for comment generation.

However, the sampling operation in Equation~\ref{eq:sampling} is not differentiable. In order to train the reader-aware saliency detection component in an end-to-end manner, we apply Gumbel-Softmax distribution as a surrogate of Bernoulli distribution for each word selection gate \cite{xue2019not}. Specifically, the selection gate produces a two-element one-hot vector as follows:
\begin{equation}
\setlength{\abovedisplayskip}{3pt}
\setlength{\belowdisplayskip}{3pt}
  \begin{array}{c}
    \mathbf{g}_{i}=\text {one\_hot}\left(\arg \max _{j} p_{i, j}, j=0,1\right)\\
    p_{i, 0}=1-\beta_i, p_{i, 1}=\beta_i
    \end{array}
\end{equation}
we use the Gumbel-Softmax distribution to approximate to the one-hot vector $\mathbf{g}_i$:
\begin{eqnarray}
  \begin{array}{c}
    \hat{\mathbf{g}}_{i}=\left[\hat{p}_{i, j}\right]_{j=0,1} \\
    \hat{p}_{i, j}=\frac{\exp \left(\left(\log \left(p_{i, j}\right)+\epsilon_{j}\right) / \tau\right)}{\sum_{j\prime=0}^{1} \exp \left(\left(\log \left(p_{i, j\prime}\right)+\epsilon_{j\prime}\right) / \tau\right)},
    \end{array}
\end{eqnarray}
where $\epsilon_{j}$ is a random sample from Gumbel(0, 1). When temperature $\tau$ approaches 0, Gumbel-Softmax distribution approaches one-hot.  And now we use $g_i=\hat{\mathbf{g}}_{i,0}$ instead of Equation \ref{eq:sampling}. Via this approximation, we can train the component end-to-end with other modules. In order to encourage the saliency detection component to turn off more gates and select less words, a $l_1$ norm term over all gates is added to the loss function as follows:
\begin{equation}
\setlength{\abovedisplayskip}{3pt}
\setlength{\belowdisplayskip}{3pt}
    \mathcal{L}_{sal}=\frac{\|\mathcal{G}\|_{1}}{|X|}=\frac{\sum_i g_i}{|X|} .
\end{equation}

\subsection{Diversified Comment Generation}
Given the learned $K$ reader-aware topic vectors, we need to select an appropriate topic for current article to guide the comment generation. Therefore, a two-layers MLP with $\operatorname{softmax}$ activation function is used to predict the selection probability of each topic as follows:
\begin{equation}
\setlength{\abovedisplayskip}{3pt}
\setlength{\belowdisplayskip}{3pt}
  p(c|X) = MLP(\mathbf{h}^e).
\end{equation}
During training, the true topic distribution $q(c|\mathbf{z})$ (Equation \ref{encode_process}) is available and is used to compute a weighted topic representation  by:
\begin{equation}
\setlength{\abovedisplayskip}{3pt}
\setlength{\belowdisplayskip}{3pt}
  \tilde{\boldsymbol{\mu}} = \sum_{c}^{K}q(c|\mathbf{z}) \odot \boldsymbol{\mu_c}.
\end{equation}
After getting the topic vector $\tilde{\boldsymbol{\mu}}$ and the selected saliency information $\tilde{\mathbf{h}}^e _ { t }$, we update the hidden vector of the backbone decoder in Section~\ref{encattdec} 2.2 as follows:
\begin{equation}
\setlength{\abovedisplayskip}{3pt}
\setlength{\belowdisplayskip}{3pt}
  \tilde{\mathbf{h}}_t = \operatorname{tanh}(W_c[\mathbf{h}_t;\tilde{\mathbf{h}}_t^e;\tilde{\boldsymbol{\mu}}]).
\end{equation}
Then $\tilde{\mathbf{h}}_t$ is used to predict next word as Equation \ref{predict}.

In the inference stage, $p(c|X)$ is used to get the topic representation. So in order to learn $p(c|X)$ during the training stage, a KL term $\mathcal{L}_{top}=D_{KL}(q(c|\mathbf{z})||p(c|X))$ is added into the final loss function.

\subsection{Learning}
Finally, considering all components the loss function of the whole comment generation  framework is as follows:
\begin{equation}\label{eq:final_loss}
\setlength{\abovedisplayskip}{3pt}
\setlength{\belowdisplayskip}{3pt}
    \mathcal{L}={\lambda_1\mathcal{L}_\mathrm{ELBO}} + \lambda_2 \mathcal{L}_{sal} + \lambda_3\mathcal{L}_{ce} +  \lambda_4 \mathcal{L}_{top}.
\end{equation}
where $\lambda_1, \lambda_2, \lambda_3$, and $\lambda_4$ are hyperparameters to make a trade-off among different components. Then we jointly train all components according to Equation \ref{eq:final_loss}.

\section{Experimental Settings}

\subsection{Datasets}
\noindent\textbf{Tencent Corpus} is a Chinese dataset published in \cite{qin-etal-2018-automatic}. The dataset is built from Tencent News\footnote{https://news.qq.com/} and each data item contains a news article and the corresponding comments. Each article is made up of a title and a body. All text is tokenized by a Chinese word segmenter JieBa\footnote{https://github.com/fxsjy/jieba}. The average lengths of news titles, news bodies, and comments are 15 words, 554 words, and 17 words respectively.

\noindent\textbf{Yahoo! News Corpus} is an English dataset published in \cite{yang-etal-2019-read}, which is built from Yahoo! News\footnote{https://news.yahoo.com/}. Text is tokenized by Stanford CoreNLP \cite{manning2014stanford}. The average lengths of news titles, news bodies, and comments are 12, 578, and 32 respectively.

\noindent\textbf{NetEase News Corpus} is also a Chinese dataset crawled from NetEase News\footnote{https://news.163.com/} and used in \cite{zheng2017automatic}.
%The original dataset used in \cite{zheng2017automatic} contains a small number of news articles and a large number of comments per article.
We process the raw data according to the processing methods used in the first two datasets \cite{qin-etal-2018-automatic,yang-etal-2019-read}.
%Specifically, we use JieBa to tokenize all text in the data. Then we filter out news articles shorter than 30 words in the body and comments shorter than 10 words or longer than 100 words. Next, news articles with less than 5 comments are removed. If the number of comments of an article exceeds 30, we only keep top 30 comments with the most upvotes. After the pre-processing, the dataset is randomly divided into  a training set, a validation set, and a test set.
On average, news titles, news bodies, and comments contain 12, 682, and 23 words respectively.

%In our experiment, we keep top 5 comments for each article in Tencent training set and top 8 comments of each article in Yahoo training set.
Table \ref{datasets} summarizes the statistics of the three datasets.

\begin{table}[!ht]
\renewcommand\arraystretch{0.8} 
\resizebox{1\columnwidth}{!}{
\begin{tabular}{@{}lrccc@{}}
\toprule
        &                       & Train   & Dev   & Test  \\ \midrule
\multirow{2}{*}{Tencent}        & \# News               & 191,502 & 5,000 & 1,610 \\
        & Avg. \# Cmts per News & 5      & 27    & 27    \\
        \midrule
\multirow{2}{*}{Yahoo}        & \# News               & 152,355 & 5,000 & 3,160 \\
        & Avg. \# Cmts per News & 7.7    & 20.5  & 20.5  \\
        \midrule
\multirow{2}{*}{NetEase}        & \# News               & 75,287        & 5,000      & 2,500      \\
        & Avg. \# Cmts per News &  22.7       & 22.5     &   22.5     \\ 
        \bottomrule
\end{tabular}
}
\caption{Statistics of the three datasets.}
\label{datasets}
\vspace{-5mm}
\end{table}

\subsection{Baseline Models}
The following models are selected as baselines:

\textbf{Seq2seq} \cite{qin-etal-2018-automatic}: this model follows the framework of seq2seq model with attention. We use two kinds of input, the title(\textbf{T}) and the title together with the content (\textbf{TC}).

\textbf{GANN} \cite{zheng2017automatic}: the author proposes a gated attention neural network, which is similar to \textbf{Seq2seq-T} and adds a gate layer between encoder and decoder.

\textbf{Self-attention} \cite{chen2018best}: this model also follows seq2seq framework and use multi-layer self multi-head attention as the encoder and a RNN decoder with attention is applied. We follow the setting of  \citeauthor{li-etal-2019-coherent}\shortcite{li-etal-2019-coherent} and use the bag of words as input. Specifically the words with top 600 term frequency are as the input. 

\textbf{CVAE} \cite{zhao2017learning}: this model uses conditional VAE to improve the diversity of neural dialog. We use this model as a baseline for evaluating the diversity of comments.

\subsection{Evaluation Metrics}

\noindent\textbf{Automatic Evaluation}
Following \citet{qin-etal-2018-automatic}, we use ROUGE \cite{lin-2004-rouge}, CIDEr \cite{vedantam2015cider},  and METEOR \cite{banerjee-lavie-2005-meteor} as metrics to evaluate the performance of different models. A popular NLG evaluation tool nlg-eval\footnote{https://github.com/Maluuba/nlg-eval} is used to compute these metrics. Besides the overlapping based metrics, we use Distinct \cite{li2016persona} to evaluate the diversity of comments.
Distinct-n measures the percentage of distinct n-grams in all generated results. 
M-Distinct-n measures the ability to generate multiple diverse comments for the same test article. For computing M-Distinct, 5 generated comments for each test article are used. For Seq2seq-T, Seq2seq-TC, GANN and Self-attention, top 5 comments from beam search with a size of 5 are used. For CVAE, we decode 5 times by sampling on latent variable to get 5 comments. For our method, we decode 5 times with one of top 5 predicted topics to get 5 comments.

\noindent\textbf{Human Evaluation}. Following \citet{qin-etal-2018-automatic}, we also evaluate our method by human evaluation. Given titles and bodies of news articles, raters are asked to rate the comments on three dimensions: \textbf{Relevance}, \textbf{Informativeness},  and \textbf{Fluency}. \textbf{Relevance} measures whether the comment is about the main story of the news, one side part of the news, or irrelevant to the news. 
% \textbf{Informativeness} evaluates how much concrete information the comment contains. It measures whether the comment involves a specific aspect of some character or event, or is a general comment that can be the answer to many news. 
\textbf{Informativeness} evaluates how much concrete information the comment contains. It measures whether the comment involves a specific aspect of some character or event.
\textbf{Fluency} evaluates whether the sentence is fluent. It mainly measures whether the sentence follows the grammar. The score of each aspect ranges from 1 to 5. In our experiment, we randomly sample 100 articles from the test set for each dataset and ask three raters to judge the quality of the comments given by different models. For every article, comments from all models are pooled, randomly shuffled, and presented to the raters. 
% Each comment is judged by the three raters on above three dimensions.

\begin{table*}[!t]
\renewcommand\arraystretch{0.8}
	\centering
	\resizebox{1.9\columnwidth}{!}{
\begin{tabular}{@{}lrccccccc@{}}
\toprule

Dataset                  & Models         & ROUGE\_L & CIDEr & METEOR & Distinct-3 & Distinct-4 & M-Distinct-3 & M-Distinct-4 \\ \midrule
\multirow{6}{*}{Tencent} & Seq2seq-T      & 0.261 & 0.015 & 0.076 & 0.088  & 0.079  & 0.046 & 0.051 \\
                         & Seq2seq-TC     & 0.280 & 0.021 & 0.088 & 0.121  & 0.122  & 0.045 & 0.054 \\
                         & GANN           & 0.267 & 0.017 & 0.081 & 0.087  & 0.081  & 0.040 & 0.046 \\
                         & Self-attention & 0.280 & 0.019 & 0.092 & 0.117  & 0.121  & 0.043 & 0.051 \\
                         & CVAE           & 0.281 & 0.021 & 0.094 & 0.135  & 0.137  & 0.041 & 0.044 \\
                         & Ours            & \textbf{0.289} & \textbf{0.024} & \textbf{0.107} & \textbf{0.176}  & \textbf{0.196}  & \textbf{0.092} & \textbf{0.112} \\
                         \midrule
\multirow{6}{*}{Yahoo}   & Seq2seq-T      & 0.299    & 0.031 & 0.105  & 0.137      & 0.168      & 0.044      & 0.063      \\
                         & Seq2seq-TC     & 0.308    & \textbf{0.037} & 0.106  & \textbf{0.179}      & 0.217      & 0.056      & 0.078      \\
                         & GANN           & 0.301    & 0.029 & 0.104  & 0.116      & 0.148      & 0.034      & 0.049      \\
                         & Self-attention & 0.296    & 0.025 & 0.096  & 0.150      & 0.181      & 0.049      & 0.068      \\
                         & CVAE           & 0.300    & 0.031 & 0.107  & 0.159      & 0.192      & 0.049      & 0.069      \\
                         & Ours           & \textbf{0.309}    & 0.033 & \textbf{0.111}  & 0.169      & \textbf{0.220}      & \textbf{0.068}      & \textbf{0.097}      \\
                         \midrule
\multirow{6}{*}{NetEase}     & Seq2seq-T      & 0.263 & 0.025 & 0.105 & 0.149  & 0.169  & 0.046 & 0.056 \\
                         & Seq2seq-TC     & 0.268 & \textbf{0.035} & 0.108 & 0.178  & 0.203  & 0.053 & 0.064 \\
                         & GANN           & 0.258 & 0.022 & 0.105 & 0.129  & 0.146  & 0.042 & 0.051 \\
                         & Self-attention & 0.265 & 0.034 & 0.110 & 0.174  & 0.204  & 0.053 & 0.067 \\
                         & CVAE           & 0.261 & 0.026 & 0.106 & 0.120  & 0.135  & 0.041 & 0.049 \\
                         & Ours            & \textbf{0.269} & 0.034 & \textbf{0.111} & \textbf{0.189}  & \textbf{0.225}  & \textbf{0.081} & \textbf{0.103} \\
                         \bottomrule
\end{tabular}
}
	\caption{Automatic evaluation results on three datasets}
	\label{overallresult}
	\vspace{-5mm}
\end{table*}

\begin{table}[!t]
\renewcommand\arraystretch{0.8}
\centering
\resizebox{0.98\columnwidth}{!}{
\begin{tabular}{@{}lrcccc@{}}
\toprule
Dataset                  & Models         & Relevance     & Informativeness & Fluency       & Total         \\ \midrule
\multirow{4}{*}{Tencent} & Seq2seq-TC     & 1.22          & 1.11            & \textbf{3.52} & 1.95          \\
                         & Self-attention & 1.48          & 1.41            & \textbf{3.52} & 2.14          \\
                         & CVAE           & 1.58          & 1.44            & 3.47          & 2.16          \\
                         & Ours           & \textbf{2.02} & \textbf{1.84}   & 3.49          & \textbf{2.45} \\ \midrule
\multirow{4}{*}{Yahoo}   & Seq2seq-TC     & 1.70          & 1.70            & 3.77          & 2.39          \\
                         & Self-attention & 1.71          & 1.72            & \textbf{3.84} & 2.42          \\
                         & CVAE           & 1.63          & 1.65            & 3.79          & 2.36          \\
                         & Ours           & \textbf{2.00} & \textbf{2.01}   & 3.71          & \textbf{2.57} \\ \midrule
\multirow{4}{*}{NetEase}     & Seq2seq-TC     & 1.97          & 1.99            & 4.03          & 2.66          \\
                         & Self-attention & 1.90          & 1.96            & 4.02          & 2.63          \\
                         & CVAE           & 1.50          & 1.53            & \textbf{4.25} & 2.42          \\ 
                         & Ours           & \textbf{2.10} & \textbf{2.15}   & 4.05          & \textbf{2.76} \\
\bottomrule
\end{tabular}
}
    \caption{Human evaluation results on three datasets}
	\label{human_result_ten}
	\vspace{-2mm}
\end{table}

% \begin{table}[!htb]
% \resizebox{1\columnwidth}{!}{
% \begin{tabular}{@{}rcccc@{}}
% \toprule
% Models         & Relevance & Informativeness & Fluency & Total \\ \midrule
% Seq2seq-TC     &           &                 &         &       \\
% Self-attention &           &                 &         &       \\
% CVAE           &           &                 &         &       \\ 
% Ours           &           &                 &         &       \\ 
% \bottomrule
% \end{tabular}
% }
%     \caption{Human evaluation results on Tencent dataset}
% 	\label{human_result_ten}
% \end{table}

% \begin{table}[!htb]
% \resizebox{1\columnwidth}{!}{
% \begin{tabular}{@{}rcccc@{}}
% \toprule
% Models         & Relevance & Informativeness & Fluency & Total \\ \midrule
% Seq2seq-TC     &           &                 &         &       \\
% Self-attention &           &                 &         &       \\
% CVAE           &           &                 &         &       \\ 
% Ours           &           &                 &         &       \\ 
% \bottomrule
% \end{tabular}
% }
%     \caption{Human evaluation results on NetEase dataset}
% 	\label{human_result_NetEase}
% \end{table}

\begin{table}[!t]
\renewcommand\arraystretch{0.8}
\centering
\resizebox{0.98\columnwidth}{!}{
\begin{tabular}{rcccc}
\toprule
Metrics               & Distinct-3 & Distinct-4 & M-Distinct-3 & M-Distinct-4 \\ \midrule
No Topic Modeling     & 0.142      & 0.151      & 0.050        & 0.060        \\
No Saliency Detection & 0.173      & 0.188      & 0.087        & 0.104        \\
Full Model            & 0.176      & 0.196      & 0.092        & 0.112       \\
\bottomrule
\end{tabular}
}
    \caption{Model ablation results on Tencent dataset}
	\label{ablationresult_ten}
	\vspace{-6mm}
\end{table}
% \begin{table}[!htb]
% \resizebox{1\columnwidth}{!}{
% \begin{tabular}{@{}rccc@{}}
% \toprule
% Metrics      & No Topic Modeling & No Saliency Detection & Full Model \\ \midrule
% Distinct-3   & 0.175             & 0.184                 & 0.189      \\
% Distinct-4   & 0.201             & 0.221                 & 0.225      \\
% M-Distinct-3 & 0.054             & 0.077                 & 0.081      \\ 
% M-Distinct-4 & 0.067             & 0.098                 & 0.103      \\ \bottomrule
% \end{tabular}
% }
%     \caption{Model ablation results on NetEase dataset}
% 	\label{ablationresult_NetEase}
% \end{table}

\subsection{Implementation Details}
For each dataset, we use a vocabulary with the top 30k frequent words in the entire data. We limit maximum lengths of news titles, news bodies and comments to 30, 600 and 50 respectively. The part exceeding the maximum length is truncated. The embedding size is set to 256. The word embedding are shared between encoder and decoder. For RNN based encoder, we use a two-layer BiLSTM with hidden size 128. We use a two-layer LSTM with hidden size 256 as decoder. For self multi-head attention encoder, we use 4 heads and two layers. For CVAE and our topic modeling component, we set the size of latent variable to 64. For our method, $\lambda_1$ and $\lambda_3$ are set to 1, $\lambda_2$ and $\lambda_4$ are set to $0.5 \times 10^{-3}$ and 0.2 respectively. We choose topic number $K$ from set [10, 100, 1000], and we set $K=100$ for Tencent dataset and $K=1000$ for other two datasets. 
% More details are included in the appendix. 
The dropout layer is inserted after LSTM layers of decoder and the dropout rate is set to 0.1 for regularization. The batch size is set to 128. We train the model using Adam \cite{kingma2014adam} with learning rate 0.0005. We also clamp gradient values into the range $[-8.0,8.0]$ to avoid the exploding gradient problem \cite{pascanu2013difficulty}. In decoding, top 1 comment from beam search with a size of 5 is selected for evaluation.

\section{Experimental Results and Discussions}

\subsection{Automatic and Human Evaluation} 
% \subsection{Automatic Evaluation} 
Automatic evaluation results on three datasets are shown in Table \ref{overallresult}. On most automatic metrics, our method outperforms baseline methods. Compared with Seq2seq-TC, Seq2seq-T and GANN perform worse in all metrics. This indicates that news bodies are important for generating better comments. The results of Self-attention and CVAE are not stable. Compared with Seq2seq-TC, Self-attention performs worse in Yahoo dataset and close in other datasets. CVAE performs better in Tencent dataset and worse in other dastasets compared with Seq2seq-TC. Compared with other methods, our method improves Distinct-4 and M-Distinct scores significantly. This demonstrates that our method can generate diversified comments according to different topics and salient information. While different comments of one article of Seq2seq-T, Seq2seq-TC, GANN and Self-attention come from the same beam, the M-Distinct scores of these methods are lower than ours. Although CVAE can generate different comments for one article by sampling on a latent variable, it obtains a worse M-Distinct score than ours. This demonstrates that the semantic change of generated comments is small when sampling on a latent variable. Our method generates comments by selecting different topic representation vectors and salient information of news, thus has a higher M-Distinct score.

% \subsection{Human Evaluation} 
Table \ref{human_result_ten} reports human evaluation results in three datasets. Because Seq2seq-T and GANN are not using news bodies and perform worse in automatic metrics, we compare our method with other methods. Our method achieves the best Total scores in three datasets. Specifically, our method mainly improves scores on Relevance and Informativeness. This shows that our method can generate more relevant and informative comments by utilizing reader-aware topic modeling and saliency information detection. However, our method performs worse in term of Fluency. We find that baselines tend to generate more generic responses, such as ``Me too.'', resulting in higher Fluency scores.

\subsection{Ablation Study} 
We conduct ablation study to evaluate the affection of each component and show the results in Table \ref{ablationresult_ten}. We compare our full model with two variants: (1) No Topic Modeling: the reader-aware topic modeling component is removed; (2) No Saliency Detection: the reader-aware saliency detection is removed. We can see that our full model obtains the best performance and two components contribute to the performance. No topic modeling drops a lot in Distinct and M-Distinct. This shows that the reader-aware topic modeling component is important for generating diversified comments. With saliency detection, the performance gets better and this  indicates that it is useful to detect important information of news for generating diversified comments.

\begin{figure}[!t]
	\centering
	\includegraphics[width=7cm,height=4.8cm]{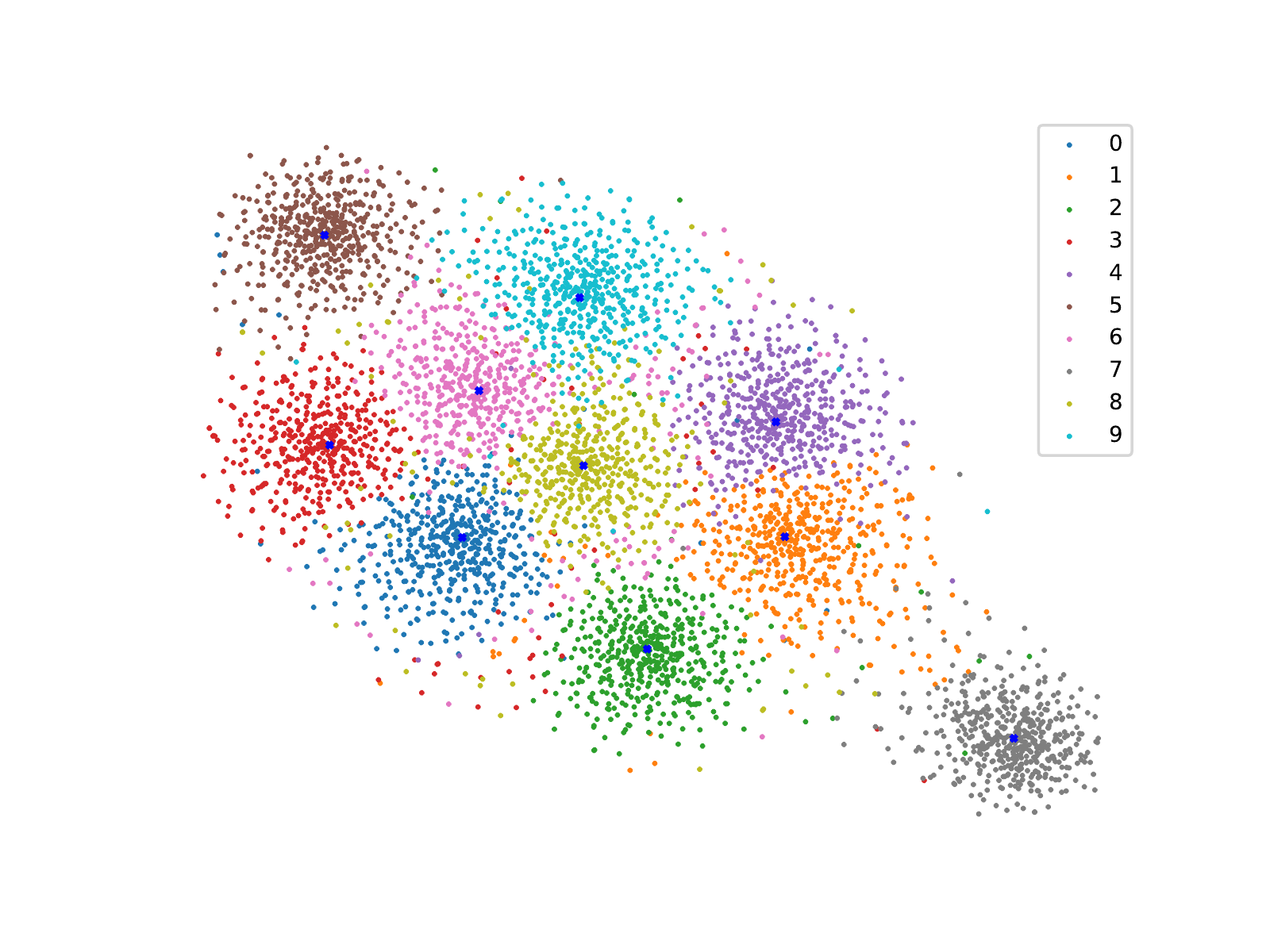}
	\caption{The latent semantic vectors of sampled comments and corresponding topic vectors.}
	\label{cluster}
	\vspace{-3mm}
\end{figure}

\begin{table}[!t]
\renewcommand\arraystretch{0.8}
\centering
% \resizebox{1\columnwidth}{!}{
\begin{tabular}{p{0.98\columnwidth}} 
\toprule
\small \textbf{Topic 1:} \songti{不错, 有点, 真的, 太, 挺, 比较, 应该, 确实, 好像, 其实 } (nice, a little, really, too, quite, comparatively, should, indeed, like, actually)   \\  
% \small \songti{不错说得好} \\
% \small \songti{哎有点小失望} \\
\midrule
\small \textbf{Topic 22:} \songti{恶心, 丑, 可爱, 真, 不要脸, 太, 讨厌, 难看, 脸, 臭 } (disgusting, ugly, cute, real, shameful, too, nasty, ugly, face, stink)
       \\ \midrule
\small \textbf{Topic 37:} \songti{好看, 漂亮, 挺, 不错, 身材, 演技, 颜值, 性感, 长得, 很漂亮}  (good-looking, beautiful, quite, nice, body, acting, face, sexy, look like, very beautiful)
   \\  \midrule
\small \textbf{Topic 62:} \songti{好吃, 东西, 喝, 吃, 味道, 鱼, 肉, 水, 菜, 不吃 } (delicious, food, drink, eat, taste, fish, meat, water, dish, don't eat) 
\\ \midrule
\small \textbf{Topic 99:} \songti{穿, 腿, 长, 衣服, 眼睛, 好看, 脸, 胖, 瘦, 高} (wear, leg, long, clothes, eye, good-looking, face, fat, thin, tall)
    \\ 
\bottomrule
\end{tabular}
% }

    \caption{The top 10 frequent words of some topics}
	\label{topic_words}
	\vspace{-7mm}
\end{table}

\subsection{Analysis of Learned Latent Topics}  
In order to evaluate the reader-aware topic modeling component, we visualize the learned latent semantic vectors of comments. To this end, we use t-SNE \cite{maaten2008visualizing} to reduce the dimensionality of the latent vector $\textbf{z}$ to 2. The one with the highest probability in topic distribution $q(c|\textbf{z})$ is used as the topic of a comment. We first randomly select 10 topics from 100 topics in Tencent dataset and then plot 5000 sampled comments belonging to these 10 topics in Figure \ref{cluster}. Points with different colors belong to different topics. In addition, we plot topic vectors of these 10 topics. We can see that these latent vectors of comments are grouped into several clusters and distributed around corresponding topic vectors. This shows that reader-aware topic modeling component can effectively cluster comments and topic vectors can be used to represent topics well. What's more, we collect comments on each topic to observe what each topic is talking about. The top 10 frequent words (removing stop words) of some topics are shown in Table \ref{topic_words}. We can see that Topic 1 is talking about intensity of emotional expression, such as ``a little'', ``really'', and ``too''. Appearance and Food are discussed in Topic 37 and Topic 62 respectively.

\subsection{Case Study} 
In order to further understand our model, we compare comments generated by different models in Table \ref{gen_case}. 
The news article is about the dressing of a female star. Seq2seq-TC produces a general comment while Self-attention produces an informative comment. However, they can not produce multiple comments for one article. CVAE can achieve this by sampling on a latent variable, but it produces same comments for different samples. Compared to CVAE, our model generates more relevant and diverse comments according to different topics. For example, ``It is good-looking in any clothes'' comments on the main story of news and mentions detail of news, ``wearing clothes''.
What's more, comparing the generated comment conditioning on a specific topic with corresponding topic words in Table \ref{topic_words}, we find that the generated comments are consistent with the semantics of the corresponding topics. 
% This shows that we can control the semantics of the generated comments by selecting different topics.

\begin{table}[!t]
\renewcommand\arraystretch{0.5}
\centering
\begin{tabular}{ll} 
\toprule
\multicolumn{2}{p{0.95\columnwidth}}{\small \textbf{Title:} \songti{蒋欣终于穿对衣服！尤其这开叉裙，显瘦20斤！微胖界穿搭典范！} (Jiang Xin is finally wearing the right clothes! Especially this open skirt, which is 20 pounds slimmer! Micro-fat dress code!)} \\ \midrule
\multicolumn{2}{p{0.95\columnwidth}}{\small \textbf{Body (truncated):}\songti{中国全民追星热的当下,明星的一举一动,以及穿着服饰,都极大受到粉丝的追捧。事实上,每位女明星都有自己的长处、优点,善意看待就好噢。蒋欣呀蒋欣,你这样穿确实很显瘦的说,着实的好看又吸睛,难怪人人都说,瘦瘦瘦也可以凹出美腻感的节奏有么有,学会这样穿说不定你也可以一路美美美的节奏。} (At the moment when China's people are star-fighting, the celebrity's every move, as well as wearing clothing, are greatly sought after by fans. In fact, each female star has its own strengths and advantages, just look at it in good faith. Jiang Xin, Jiang Xin, you are indeed very thin when you wear it.It is really beautiful and eye-catching. No wonder everyone says that there is a rhythm of thinness.You can learn to wear it like this. You can have a beautiful rhythm.)}                                                             \\
\midrule
\multicolumn{2}{p{0.95\columnwidth}}{\small \textbf{Seq2seq-TC:} \songti{我也是} (Me too.) } \\
\multicolumn{2}{p{0.95\columnwidth}}{\small \textbf{Self-attention:} \songti{喜欢蒋欣} (Like Jiang Xin.)  } \\
\midrule
\multirow{2}{*}{\small \textbf{CVAE}} & \small \textbf{1:} \songti{我喜欢蒋欣} (I like Jiang   Xin.)                            \\
                      & \small \textbf{2:} \songti{我喜欢蒋欣} (I like Jiang Xin.)                              \\
                      \midrule
\multirow{5}{*}{\small \textbf{Ours}} & \multicolumn{1}{p{0.8\columnwidth}}{\small \textbf{Topic 99:} \songti{穿什么衣服都好看} (It   is good-looking in any clothes.)} \\
                      & \multicolumn{1}{p{0.8\columnwidth}}{\small \textbf{Topic 22:} \songti{好可爱} (So cute.)   }                               \\
                      & \multicolumn{1}{p{0.8\columnwidth}}{\small \textbf{Topic 37:} \songti{挺好看的}(It is pretty beautiful.) }                 \\
                      & \multicolumn{1}{p{0.8\columnwidth}}{\small \textbf{Topic \,\,\,1:}  \songti{不错不错} (It is nice.)      }                         \\
                      & \multicolumn{1}{p{0.8\columnwidth}}{\small \textbf{Topic 62:} \songti{胖了} (Gain weight.)        }                      \\
                      \bottomrule
\end{tabular}
    \caption{A Case from Tencent News dataset}
	\label{gen_case}
	\vspace{-7mm}
\end{table}

\section{Related Work}
Automatic comment generation is proposed by \citeauthor{zheng2017automatic}\shortcite{zheng2017automatic} and \citeauthor{qin-etal-2018-automatic}\shortcite{qin-etal-2018-automatic}. The former proposed to generate one comment only based on the news title while the latter extended the work to generate a comment jointly considering the news title and body content. These two methods adopted seq2seq~\cite{sutskever2014sequence} framework and conducted saliency detection directly via attention modeling. Recently, \citeauthor{li-etal-2019-coherent}\shortcite{li-etal-2019-coherent} extracted keywords as saliency information and \citeauthor{yang-etal-2019-read}\shortcite{yang-etal-2019-read} proposed a reading network to select important spans from the news article. \citeauthor{huang2020generating}\shortcite{huang2020generating} employed the LDA \cite{blei2003latent} topic model to conduct information mining from the content. All these works concern the saliency information extraction. However, they neglect various reader-aware factors implied in the comments. Our method simultaneously considers these two aspects and utilizes two novel components to generate diversified comments.  

\section{Conclusion}
We propose a reader-aware topic modeling and saliency detection framework to enhance the quality of generated comments. We design a variational generative clustering algorithm for topic mining from reader comments. We introduce Bernoulli distribution estimating on news content to select saliency information. Results show that our framework outperforms existing baseline methods in terms of automatic metrics and human evaluation.

\clearpage
\section{Ethics Statement}

We are fully aware of the new ethics policy posted in the AAAI 2021 CFP page and we seriously honor the AAAI Publications Ethics and Malpractice Statement, as well as the AAAI Code of Professional Conduct. And along the whole process of our research project, we carefully think about them. Here we elaborate on the ethical impact of this task and our method.

Automatic comment generation aims to generate comments for news articles. This task has many potential applications.
First, researchers have been working to develop a more intelligent chatbot (such as XiaoIce \cite{shum2018eliza,zhou2020design}), which can not only chat with people, but also write poems, sing songs and so on. The application of this task is to give the chatbot the ability to comment on articles \cite{qin-etal-2018-automatic,yang-etal-2019-read} to enable in-depth, content-rich conversations with users based on articles users are interested in. 
Second, it can be used for online discussion forums to increase user engagement and foster online communities by generating some enlightening comments to attract users to give their own comments. 
Third, we can build a comment writing assistant which generates some candidate comments for users \cite{zheng2017automatic}. Users could select one and refine it, which makes the procedure more user-friendly.
Therefore, this task is novel and meaningful.

We are aware that numerous uses of these techniques can pose ethical issues. For example, there is a risk that people and organizations could use these techniques at scale to feign comments coming from people for purposes of political manipulation or persuasion \cite{yang-etal-2019-read}. Therefore, in order to avoid potential risks, best practices will be necessary for guiding applications and we need to supervise all aspects when deploying such a system. 
First, we suggest that market regulators must monitor organizations or individuals that provide such services to a large number of users.
Second, we suggest limiting the domain of such systems and excluding the political domain. And some post-processing techniques are need to filter some sensitive comments. 
Third, we suggest limiting the number of system calls in a short period of time to prevent massive abuse.
We believe that reasonable guidance and supervision can largely avoid these risks.

On the other hand, we have to mention that some typical tasks also have potential risks. For example, the technology of dialogue generation \cite{zhang-etal-2020-dialogpt} can be used to disguise as a normal person to deceive people who chat with it, and achieve a certain purpose. The technology of face generation \cite{karras2018progressive} can be used to disguise as the face of target people, so as to deceive the face recognition system. However, there are still many researchers working on these tasks for the positive uses of these tasks. Therefore, everything has two sides and we should treat it dialectically.
% However, the positive uses of these tasks are obvious. Dialogue generation can be used to comfort users who are not in a good mood; face generation can be used for virtual human synthesis to help remote virtual conference, and so on.

In addition, we believe that the study of this technology is important for us to better understand the defects of this technology, which helps us to detect spam comments and combat this behavior. For example, \citeauthor{zellers2019defending}\shortcite{zellers2019defending} found that the best defense against fake news turns out to be a strong fake news generator.

\section{Acknowledgments}
This research is supported by National Natural Science Foundation of China (Grant No. 61773229 and 6201101015), Tencent AI Lab Rhino-Bird Focused Research Program (No. JR202032), Shenzhen Giiso Information Technology Co. Ltd., the Basic Research Fund of Shenzhen City (Grand No. JCYJ20190813165003837), and Overseas Cooperation Research Fund of Graduate School at Shenzhen, Tsinghua University (Grant No. HW2018002).

\bibliography{emnlp2020}

\end{document}